\documentclass[sigconf]{acmart}

\usepackage{amssymb}
\usepackage{makecell}
\usepackage{multirow}
\usepackage{cuted} 
\usepackage{algorithm}
\usepackage{hyperref}
\usepackage{siunitx}
\usepackage[most]{tcolorbox}
\tcbset{
  promptbox/.style={
    colback=gray!5,
    colframe=gray!60,
    boxrule=0.4pt,
    arc=2pt,
    left=6pt, right=6pt, top=4pt, bottom=4pt,
    enhanced jigsaw,
  }
}
\usepackage{pifont}
\AtBeginDocument{%
  }

\setcopyright{acmlicensed}
\copyrightyear{2026}
\acmYear{2026}
\setcopyright{cc}
\setcctype{by}
\acmConference[MM '26]{Proceedings of the 34th ACM International Conference on Multimedia}{November 10--14, 2026}{Rio de Janeiro, Brazil}
\acmBooktitle{Proceedings of the 34th ACM International Conference on Multimedia (MM '26), November 10--14, 2026, Rio de Janeiro, Brazil}
\acmDOI{10.1145/3767308.3837700}
\acmISBN{979-8-4007-2213-4/2026/11}

\begin{document}

\title{Team MSU GenText-Forensics Challenge 2026 Technical Report}
\settopmatter{printfolios=false}
\author{Kirill Koltsov}
\email{kirill.e.koltsov@mail.ru}
\affiliation{%
  \institution{Lomonosov Moscow State University}
  \city{Moscow}
  \country{Russia}}

\author{Aleksandr Gushchin}
\email{alexander.gushchin@graphics.cs.msu.ru}
\affiliation{%
  \institution{MSU Institute for Artificial Intelligence}
  \city{Moscow}
  \country{Russia}}

\author{Dmitriy Vatolin}
\email{dmitriy@graphics.cs.msu.ru}
\affiliation{%
  \institution{MSU Institute for Artificial Intelligence}
  \city{Moscow}
  \country{Russia}}

\author{Anastasia Antsiferova}
\email{aantsiferova@graphics.cs.msu.ru}
\affiliation{%
  \institution{MSU Institute for Artificial Intelligence}
  \city{Moscow}
  \country{Russia}}

\renewcommand{\shortauthors}{Kirill Koltsov, Aleksandr Gushchin, Dmitriy Vatolin, \& Anastasia Antsiferova}

\begin{abstract}
  Document text forgery has evolved beyond simple pixel-level manipulation: modern attacks alter not only the appearance of a document but also its meaning, and increasingly target the OCR$\rightarrow$LLM pipelines that consume such documents. The ACM MM 2026 \\ GenText-Forensics challenge therefore requires systems that not only decide whether a multilingual text image is forged, but also localize the point of manipulation, identify the attack type, and produce a human-readable forensic report with supporting evidence. We present our solution, a \emph{decomposed} chain-of-thought (CoT) pipeline that combines a document tampering detector (DTD) with two Qwen3-VL-32B vision--language models, each LoRA-adapted to a distinct sub-task. DTD produces tampering probability maps that are converted into numbered candidate regions; a first model (the \emph{Filterer}) validates these regions and assigns a preliminary forgery type, while a second model (the \emph{Semantic Detective}) merges and re-grounds the surviving regions, searches for purely semantic anomalies that are invisible to pixel-level detectors, and writes the final report. Both models are trained by distilling chain-of-thought traces from a privileged Qwen3-VL-235B teacher that has access to ground-truth masks and reports. Our approach secured third place in the ACM MM 2026 GenText-Forensics challenge. We describe the data preparation, test-time augmentation, region rendering, distillation protocol, and training configuration in detail, and report ablations over detector thresholds, prompt designs, and pipeline decompositions.
\end{abstract}



\maketitle

\section{Introduction}

Text-rich document images---identity documents, invoices, contracts, medical
reports, bank statements---underpin countless administrative and financial
processes. As these processes are increasingly automated through optical
character recognition (OCR) followed by large-language-model (LLM) reasoning,
the integrity of the underlying image becomes a security concern: a forger who
alters a single digit, swaps one clean word for another, or splices in a
redaction block can make downstream pipelines act on false information while
leaving little visible trace.

The ACM MM 2026 GenText-Forensics challenge formalizes this threat. Beyond the
binary ``is this document forged?'', a system must (i)~\emph{detect} tampering
in a multilingual text image, (ii)~\emph{localize} the manipulated region,
(iii)~\emph{classify} the attack type, and (iv)~\emph{explain} the finding in a
structured forensic report. Submissions are scored by a detection metric
($S_\mathrm{Det}$), a pixel-level localization metric ($S_\mathrm{Loc}$), an
explanation-quality metric ($S_\mathrm{Exp}$), and an LLM-judged
report-quality metric ($S_\mathrm{Rep}$). The difficulty is compounded by
multilinguality and by two very different attack families: \emph{visual}
forgeries that leave low-level statistical artifacts, and \emph{semantic}
forgeries that are visually perfect yet logically inconsistent.

\paragraph{Related work.}
Two complementary lines of work address these families. On the visual side, the
Document Tampering Detector (DTD) of Qu et al.~\cite{qu2023doctamper} fuses a
Frequency Perception Head with a Multi-view Iterative Decoder to predict dense
pixel-probability maps of tampering; such detectors flag visual edits well but
are blind to semantic manipulations and produce no natural-language
explanation. We use the ForensicHub~\cite{forensichub2025} implementation of DTD
as our visual front-end. On the language side, multimodal LLMs yield
interpretable outputs: ForgeryGPT~\cite{forgerygpt2024} adds a localization
pathway to an MLLM, and document-centric systems such as
LogicLens~\cite{logiclens2025} structure the analysis as explicit multi-stage
reasoning. Our method builds directly on this multi-stage reasoning philosophy.

\paragraph{Our approach.}
We combine the strengths of both families and \emph{decompose} the language
model's workload across specialized components. Building on the multi-stage
reasoning philosophy of LogicLens~\cite{logiclens2025}, our pipeline
(i)~extracts visual artifact regions from DTD, (ii)~delegates region
validation and preliminary typing to one LoRA-adapted Qwen3-VL-32B model, and
(iii)~delegates semantic discovery and report generation to a second,
differently LoRA-adapted Qwen3-VL-32B model. Both language models are trained
by distilling chain-of-thought traces from a Qwen3-VL-235B teacher
that, during data generation only, has access to the ground-truth mask and
report. This separation lets each component be trained and tuned for one
well-defined objective; the resulting system reached third place in the
challenge. Our contributions are: (1)~a decomposed DTD$+$dual-VLM pipeline for
joint detection, localization, typing and explanation; (2)~a privileged-teacher
CoT distillation protocol that ties detector regions to ground-truth anomalies
and splits the resulting trace across two students; (3)~a practical study of
detector thresholding, test-time augmentation, region rendering, and prompt
design, with ablations isolating the contribution of each choice.

\paragraph{Code and data.}
Our code and prompt templates are released on
\href{https://github.com/koltsov-cmc/GenText-Forensics-Challenge-2026-solution-code}{GitHub}
and the accompanying data on
\href{https://huggingface.co/datasets/cmcshnik/GenText-Forensics_third_place_additional_materials}{Hugging~Face}.
The data release includes the distilled chains of thought for the training
images and, for the test set, the marked-up images together with the rendered
prompts and the technical artifacts from our runs (PP-OCRv5 OCR output and DTD
detection results).

\section{Method}

\subsection{Overview}

\begin{figure*}[t]
  \centering
    \includegraphics[width=\textwidth]{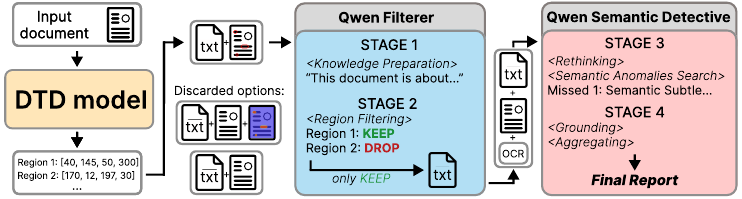}
  \caption{Overview of the decomposed chain-of-thought pipeline. Document icons labeled "txt" denote the natural-language prompts supplied to the Qwen models. These prompts are not fixed: a distinct template is used at each stage, and each template is populated dynamically with stage-specific evidence before generation. Prior to the Qwen Filterer, the prompt is augmented with the numbered DTD candidate regions (each index matching a red box on the annotated image). Prior to the Qwen Semantic Detective, the prompt is augmented with only the regions the Filterer marked KEEP, together with the full output of the OCR model—the recognized text of every word and its bounding-box coordinates in the source document—so that the model can ground its findings on observed token positions.}
  \Description{Block diagram of the five pipeline stages from input document
    image to final forensic report.}
  \label{fig:pipeline}
\end{figure*}

Our system follows the multi-stage reasoning structure of LogicLens~\cite{logiclens2025}, but replaces its single network with a third-party tampering detector and two separately specialized language models. The end-to-end flow for one document is as follows
(depicted in the Figure~\ref{fig:pipeline}):

\begin{enumerate}
  \item The document image is processed by DTD~\cite{qu2023doctamper,forensichub2025}
    (Document Tampering Detector)---an external, pre-existing detector that we
    use off-the-shelf---which searches for
    statistical deviations in the image space (font mismatches, pixelation,
    blurring, redaction blocks, kerning and baseline anomalies, and JPEG
    discontinuities). Because DTD operates on $512\times512$ inputs, the image
    is tiled into overlapping patches, each scored by the detector, and the
    per-patch probabilities are stitched back into a full-resolution map.
  \item The probability map is thresholded and converted into \\ connected-region
    bounding boxes. Each region is numbered, and a numbered red box is drawn
    on the \emph{original} image at the corresponding location.
  \item The annotated image, together with an augmented textual prompt listing the numbered regions, is passed to the \textbf{Qwen Filterer}, a LoRA-adapted~\cite{hu2022lora} Qwen3-VL-32B~\cite{qwen3vl2025}. It validates each DTD region and labels it \texttt{KEEP} or \texttt{DROP}, attaching a preliminary forgery type to the kept regions. Every \texttt{DROP} region is then discarded.
  \item The Filterer's output is post-processed and handed to the \textbf{Qwen Semantic Detective}, a second LoRA-adapted Qwen3-VL-32B. It additionally receives the full output of the PP-OCRv5 OCR engine~\cite{paddleocr}---word-level triplets \\  $(\langle\textit{text}\rangle, \langle\textit{bbox}\rangle, \langle\textit{conf}\rangle)$ plus the whole text from the document in reading order.
  \item The Semantic Detective re-evaluates regions that were marked as KEEP, merges those belonging to the same semantic span, searches for semantic anomalies that DTD cannot see, grounds every finding against the OCR coordinates, and produces the final forensic report.
\end{enumerate}

The remainder of this section describes each component in turn.

\subsection{Visual Tampering Detection with DTD}

The DTD component clearly isolates the obvious visual artifacts before any language model is invoked. DTD was introduced by Qu et al.~\cite{qu2023doctamper} for tampered-text detection in document images; it predicts a dense pixel-level tampering probability map by fusing a spatial (visual) stream with a frequency stream that captures compression and typographic irregularities. We adopt the ForensicHub~\cite{forensichub2025} re-implementation, with a ConvNeXt-Small spatial backbone and a Swin Transformer V2-Small backbone, and a DCT quantization table supplying the frequency input.

\paragraph{Training configuration.}
We fine-tune DTD on the challenge data. Inputs are resized to $512\times512$ with ImageNet normalization, and the model predicts a tampering mask only (no image-level label head). Optimization uses AdamW with a weight decay of $0.05$, a base learning rate of $1\times10^{-4}$ decaying to a minimum of $5\times10^{-7}$, and one warm-up epoch, run for $20$ epochs with mixed precision, a per-GPU batch size of $16$, and synchronous data-parallel training across eight GPUs (random seed $42$). Pixel-F1 and Image-F1 are tracked on a held-out split during training.

\paragraph{Patch construction.}
Because DTD requires $512\times512$ inputs \\ whereas challenge documents are of arbitrary size, each image is tiled into a grid of $512\times512$ patches. Patches at the right and bottom borders are \emph{not} cropped to a smaller size; instead they are taken overlapped slightly with the preceding patches, so that every output pixel is covered by a complete-resolution prediction. We reserved $8{,}000$ training images for DTD; tiling them produced a total of $53{,}751$ patches, on which DTD was trained. A further $4{,}000$ images (disjoint from the $8{,}000$) together with $500$ images set aside for local testing were tiled in the same way and used for validation during DTD training.

\paragraph{Probability thresholding.}
DTD produces a per-pixel probability map in which higher values indicate a higher likelihood of tampering. Converting this map into discrete candidate regions requires a decision threshold, which directly trades false positives against missed detections. Sweeping the threshold from $0.20$ to $0.90$ on the forged subset of our $500$ held-out test images, localization quality peaks at $0.40$ ($\mathrm{mIoU}=0.5792$, $\mathrm{mF1}=0.6855$) and degrades gradually on either side ($\mathrm{mIoU}=0.5613$ at $0.20$ and $0.5477$ at $0.90$); we therefore adopt $0.40$ for all subsequent stages. Connected components below a minimum area are discarded, and a light morphological opening is applied before box extraction.

\paragraph{Test-time augmentation.}
A systematic failure mode arises from the patch grid itself: where the tiling boundary cuts a word in two, DTD frequently raises a false alarm on the truncated fragment. To suppress these boundary artifacts we apply test-time augmentation (TTA). Each image is processed four times with different tiling origins: the original grid, the grid shifted $256$ pixels to the right, the grid shifted $256$ pixels down, and the grid shifted $256$ pixels in both directions. The four probability maps are combined conservatively (pixel-wise minimum), so that only regions flagged in \emph{all four} passes survive; a spurious detection caused by one particular cut disappears under the shifted grids and is therefore discarded. This suppresses boundary false alarms without removing genuine tampering, and yields roughly a $0.04$ increase in the final score.

\paragraph{Training data composition.}
We observed that training DTD only on forged documents yielded weaker results than training on a mixture of forged and pristine documents, most plausibly because the latter exposes the detector to more data and to a wider range of benign document content, sharpening its discrimination of genuine artifacts from ordinary layout. When training only on forged mIoU drops from 0.5792 to 0.4992.

\paragraph{Region-splitting limitations.}
When two distinct tampered regions touch, the bounding-box extraction occasionally encloses them in a single box instead of the two separate boxes. We experimented with a geometric splitting heuristic that tries to break such a merged box apart: if the flagged pixels fill only a small fraction of their bounding rectangle (a sign that the box actually spans two separate blobs with empty space between them), the box is split in two at the widest empty band---the largest run of background rows or columns inside it---and the procedure is applied again to each half. On the test split this did not improve results, and we therefore do not rely on it for the reported configuration.

\subsection{Qwen Filterer}

The Filterer's task is to remove false DTD regions and retain only those that genuinely cover tampered content. We use Qwen3-VL-32B model LoRA-adapted on specially distilled chains of thought (Section~\ref{sec:distill}).

\paragraph{Choice of visual--textual input.}
A central design question is how to present the DTD regions to the model. In every case the prompt lists the numbered DTD regions together with their pixel coordinates; what we vary is only the \emph{visual} input that accompanies this text. We compared three visual formats:

\begin{enumerate}
  \item \emph{Clean image.} The unmodified image is provided, so the numbered regions exist only as coordinates in the prompt. This works poorly: the model struggles to map the textual pixel coordinates onto the image and to localize the regions reliably.
  \item \emph{Clean image $+$ heatmap overlay.} A second image, onto which the DTD activation heatmap is projected, is provided alongside the original. This also works poorly: the model is overwhelmed by the volume of visual information and loses track of the relevant regions.
  \item \emph{Image with numbered red boxes.} The original image is annotated with numbered red bounding boxes drawn at the DTD regions, matching the numbering used in the prompt. This is the most effective format: the visual anchor lets the model navigate directly to each region. To keep the region indices legible across document scales, each index is rendered at a font size proportional to the height of its box (the desired glyph height is set to a fixed fraction of the box height in pixels). A by-product is that larger, more prominent regions naturally carry larger labels, focusing attention where artifacts are most salient.
\end{enumerate}

According to the results shown in Table~\ref{tab:choice} we adopt the third format. Alongside the annotated image, the Filterer receives a prompt that enumerates the DTD regions and states the rules by which they are to be filtered. The prompt enforces an explicit two-stage format: in \textbf{Stage~1 (Knowledge Preparation)} the model forms a global understanding of the document, following the LogicLens~\cite{logiclens2025} protocol; in \textbf{Stage~2 (Per-Region Walk-Through)} it inspects each region and assigns it a \texttt{KEEP} or \texttt{DROP} label. Each decision must be justified against the rules in the prompt---which we refined empirically during development---and each \texttt{KEEP} region is given a preliminary forgery type (Visual~Clumsy, Logical~Fraud, or Semantic~Subtle). Concrete prompt templates are released alongside the code on Hugging Face.

\paragraph{Why a language model.}
We also experimented with filtering DTD regions using purely visual classifiers (a ResNet and a U-Net). These did not yield competitive results, which we attribute to their lack of the document-level semantic understanding that a large vision--language model provides: deciding whether a flagged region is a genuine forgery often requires reading the surrounding text and reasoning about its plausibility according to the context of the document, not merely inspecting local pixels. Table~\ref{tab:filter} reports the contribution of the Filtering stage.

\begin{table}[t]
  \caption{Choice of visual–textual input, measured on all 500 held-out test images with Two-Stage pipeline}
  \label{tab:choice}
  \begin{tabular}{lc}
    \toprule
    Filtering of DTD regions & Final Score \\
    \midrule
    Clean image & $0.659$ \\
    Clean Image + heatmap overlay   & $0.751$ \\
    Image with numbered red boxes     & $\mathbf{0.794}$ \\
    \bottomrule
  \end{tabular}
\end{table}

\begin{table}[t]
  \caption{Effect of the Qwen Filterer on downstream localization, measured on
    all $500$ held-out test images (forged and pristine).}
  \label{tab:filter}
  \begin{tabular}{lcc}
    \toprule
    Filtering of DTD regions & mIoU & mF1 \\
    \midrule
    None (raw DTD regions)   & $0.6402$ & $0.7058$ \\
    ResNet / U-Net filter    & $0.6113$ & $0.6510$ \\
    Qwen Filterer (ours)     & $\mathbf{0.6672}$ & $\mathbf{0.7404}$ \\
    \bottomrule
  \end{tabular}
\end{table}

\subsection{Qwen Semantic Detective}

The Semantic Detective, the second LoRA-adapted Qwen3-VL-32B, has three responsibilities:

\begin{enumerate}
  \item \emph{Re-think the kept DTD regions:} merge regions that clearly belong to the same semantic span of text into a single anomaly.
  \item \emph{Find purely semantic anomalies:} because DTD detects only visual forgeries, semantic manipulations that are visually flawless (e.g.\ a clean word substituted for another clean word) leave no trace for the detector; the Detective must find these from the text relying on the surrounding context.
  \item \emph{Prepare the final report:} produce concise descriptions of every confirmed anomaly and synthesize them into the required report format.
\end{enumerate}

\paragraph{Inputs.}
The Filterer's output is first post-processed algorithmically: regions marked \texttt{DROP} are removed, and only \texttt{KEEP} regions are propagated. The Semantic Detective then receives the document image together with a prompt containing (i)~the Filterer's Stage~1 and Stage~2 blocks, where Stage~2 has been reduced to the \texttt{KEEP} regions only; (ii)~the complete PP-OCRv5 output as word-level triplets $(\langle\textit{text}\rangle, \langle\textit{bbox}\rangle, \langle\textit{conf}\rangle)$ for localizing individual tokens; and (iii)~the full document text in reading order.

\paragraph{Reasoning and report.}
The prompt mandates a clear structure. Inside a \verb|<think>|\,\dots\,\verb|</think>| section, the model performs \textbf{Stage~3 (Independent Semantic and Logical Search, and Rethinking)}, where it both re-examines and merges the kept regions and independently searches for semantic/logical inconsistencies, and \textbf{Stage~4 (Final Naming, Ordering, Grounding, and Bridging)}, where it finalizes each anomaly and grounds it using the OCR coordinates. The grounding step therefore anchors every reported region to concrete OCR bounding boxes rather than to free-hand estimates. \emph{Bridging} is the explicit final step that connects all of the preceding reasoning to the generation of the report: it consolidates the kept regions, the newly found semantic anomalies, their assigned types and their grounded coordinates into a single coherent thread, so that the report follows from the chain of thought rather than being produced independently of it. Inside a \verb|<report>|\,\dots\,\verb|</report>| section, the model then transcribes this bridged reasoning into the final forensic report.

This two-stage organization relieves a single model from having to do everything at once---locate visual artifacts, reason about them, and immediately write a report. By decomposing the task we can also train two specialized models, one per half of the reasoning trace. The effectiveness of this decomposition relative to single-pass alternatives is studied in Section~\ref{sec:experiments}.

\subsection{Chain-of-Thought Distillation}
\label{sec:distill}

Realizing the multi-stage CoT pipeline requires supervision in exactly the target format, which we obtain by distillation from a Qwen3-VL-235B-A22B-Instruct teacher on $4{,}000$ images.

\paragraph{Privileged teacher.}
During distillation only, the teacher is given ground-truth information the students never see. Each ground-truth (GT) mask is converted into precise tampering boxes (connected components), and every DTD region is labelled \texttt{TP} or \texttt{FP} by an intersection-over-minimum-area match against these boxes (sufficient overlap $\Rightarrow$ keep). A teacher prompt then lists, per image, the \texttt{TP}/\texttt{FP} regions, what DTD \emph{missed}, and the GT report, with an instruction to remain faithful to it; a separate variant handles pristine documents, where every DTD region is a false positive and the verdict is \textsc{authentic}.

\paragraph{Coordinate refinement.}
To make the supervision spatially precise, each GT anomaly's coordinates are refined before the report reaches the teacher: a GT box is matched against the DTD regions and OCR word boxes, preferring the DTD box (tighter localization), falling back to OCR, and otherwise keeping the original GT box. The refined coordinates are written back into the report, so the teacher reasons over the \emph{actual} DTD/OCR boxes the student can reproduce at inference time. As an upper bound, replacing every GT region by its best-matching DTD or OCR box yields $\mathrm{mIoU}=0.7187$ and $\mathrm{mF1}=0.7974$---the maximum ``data purity'' from these two sources; each match is the individual box or union with the highest IoU among all boxes overlapping the GT region ($\mathrm{IoU}>0.1$).

\paragraph{Generation and validation.}
Given the image, the labelled DTD regions, and the refined report, the teacher generates the full Stage~1--4 chain of thought in one pass, served with vLLM~\cite{kwon2023vllm} in bfloat16 (tensor-parallel over eight GPUs, greedy decoding, up to $8{,}192$ tokens, two images per prompt). Every trace is validated automatically---it must contain a \verb|<think>| and a \verb|<report>| block, expose at least four stages, reproduce the augmented GT report, and avoid a blacklist of phrases that would leak the ground truth (e.g.\ ``the mask shows'')---and failing traces are excluded, keeping the training data free of leakage and format drift.

\subsection{Chain-of-Thought Training}

Once distilled, each chain-of-thought trace is split into two parts that serve as targets for the two students:

\begin{itemize}
  \item The \textbf{first half} contains only Stage~1 and Stage~2; it is used as the target for LoRA fine-tuning of the \emph{Qwen Filterer}.
  \item The \textbf{second half} contains only Stage~3 and Stage~4 (and the final report); all \texttt{DROP} regions are algorithmically removed from Stage~3 to match inference, where only \texttt{KEEP} regions remain. This half is used as the target for the \emph{Qwen Semantic Detective}.
\end{itemize}

In other words, each CoT block is divided into two segments, and each segment trains the corresponding student. Both students start from Qwen3-VL-32B-Instruct and are trained with QLoRA~\cite{dettmers2023qlora}. We use QLoRA---rather than plain LoRA---purely for memory reasons: quantizing the $32$B base model to 4-bit NF4 with double quantization lets the frozen weights, the LoRA adapter, and the activations fit on our GPUs, while the small LoRA adapter on the attention projection matrices carries the actual adaptation in a bfloat16 compute dtype. The vision tower is kept frozen. Table~\ref{tab:lora} lists the training configuration; the two students differ only in their target segment of the CoT trace (and the Filterer additionally consumes the annotated image, whereas the Detective's stage runs without the heatmap). At inference time, both LoRA-adapted students are served with vLLM~\cite{kwon2023vllm}---the same serving stack used for the teacher during distillation.

\begin{table}[t]
  \caption{QLoRA fine-tuning configuration for the two students.}
  \label{tab:lora}
  \begin{tabular}{ll}
    \toprule
    Setting & Value \\
    \midrule
    Base model            & Qwen3-VL-32B-Instruct \\
    Quantization          & 4-bit NF4, double quant., bf16 compute \\
    LoRA rank $r$         & $16$ \\
    LoRA $\alpha$         & $32$ \\
    LoRA dropout          & $0.05$ \\
    Target modules        & \texttt{q,k,v,o\_proj} \\
    Vision tower          & frozen \\
    Learning rate         & $1\times10^{-4}$ \\
    LR schedule           & cosine, warm-up ratio $0.03$ \\
    Weight decay          & $0.0$ \\
    Optimizer             & AdamW \\
    Per-device batch size & $1$ \\
    Gradient accumulation & $8$ \\
    Max gradient norm     & $1.0$ \\
    Gradient checkpointing & enabled \\
    Precision             & bf16 (tf32 enabled) \\
    Seed                  & $42$ \\
    \bottomrule
  \end{tabular}
\end{table}

\section{Experiments}
\label{sec:experiments}

\subsection{Evaluation protocol}

The challenge score is a weighted combination of four components: a detection F1 ($S_\mathrm{Det}$), a pixel-level grounding score ($S_\mathrm{Loc}=\tfrac12\,\mathrm{mIoU}+\tfrac12\,\mathrm{mF1}$), an explanation score ($S_\mathrm{Exp}$, BERTScore-F1~\cite{zhang2020bertscore} with an XLM-RoBERTa-Large backbone), and a report-quality score ($S_\mathrm{Rep}$, the mean of factuality, reasoning and completeness ratings from a Qwen3-VL-32B LLM judge). We report all four components and the final weighted score.

\subsection{Prompt configurations}

We evaluated four prompt/pipeline configurations, each run both as a zero-shot prompt on the base Qwen3-VL-32B and as a LoRA-fine-tuned model. To isolate the effect of the configuration itself, all of them are fed the same input form---the stage prompt together with the original image annotated with numbered red boxes---and differ only in how the DTD regions are treated, whether intermediate reasoning is produced, and what the training target is:

\begin{description}
  \item[\textsc{strong-direct}.] The model emits the final report \emph{immediately, without any intermediate reasoning}, and is required to keep every DTD region: none may be dropped.
  \item[\textsc{direct}.] As \textsc{strong-direct}, the model emits the report directly without reasoning, but here it \emph{is} permitted to drop DTD regions that the prompt's rules identify as false positives.
  \item[\textsc{all-in-one}.] The full multi-stage reasoning of our pipeline, but performed by a \emph{single} Qwen that produces the entire Stage~1--4 trace and the report in one pass. It is trained on exactly the same distilled CoTs as \textsc{two-stage}, only without splitting the work across two specialized models.
  \item[\textsc{two-stage} (final, best).] Our full decomposed pipeline: a Filterer (Stage~1--2, KEEP/DROP $+$ type) followed by a Semantic Detective (Stage~3--4 $+$ report), executed by two separately LoRA-adapted models.
\end{description}

The configurations also differ in their training targets. For the two direct configurations (\textsc{strong-direct} and \textsc{direct}), which emit no reasoning, the LoRA target is the ground-truth report itself. For \textsc{all-in-one} and \textsc{two-stage}, the target is the distilled chain of thought concatenated with the ground-truth report, so the model learns to reason before it reports.

Table~\ref{tab:configs} compares these configurations. The two-stage
decomposition is our final submission and attains the best overall score.

\begin{table}[t]
  \small
  \caption{Comparison of prompt / pipeline configurations on the
    $N{=}100$-document evaluation set. ``Trained'' indicates LoRA fine-tuning
    on the corresponding targets, as opposed to zero-shot prompting of the base
    model. $S_\mathrm{Det}$, $S_\mathrm{Loc}$ and $S_\mathrm{Exp}$ lie in
    $[0,1]$; the LLM-judge column ($S_\mathrm{Rep}$) is on a $0$--$100$ scale.
    The final submission is in bold.}
  \label{tab:configs}
  \begin{tabular}{llccccc}
    \toprule
    Configuration & Trained & $S_\mathrm{Det}$ & $S_\mathrm{Loc}$ &
      $S_\mathrm{Exp}$ & $S_\mathrm{Rep}$ & Final \\
    \midrule
    \textsc{strong-direct} & No  & $\mathbf{1.00}$ & $0.55$ & $0.86$ & $61.2$ & $0.764$ \\
    \textsc{direct}        & No  & $\mathbf{1.00}$ & $0.55$ & $0.86$ & $61.6$ & $0.768$ \\
    \textsc{all-in-one}    & No  & $0.82$ & $0.22$ & $0.86$ & $51.6$ & $0.566$ \\
    \textsc{two-stage}     & No  & $0.92$ & $0.51$ & $0.86$ & $56.8$ & $0.721$ \\
    \midrule
    \textsc{strong-direct} & Yes & $0.97$ & $0.62$ & $\mathbf{0.88}$ & $68.4$ & $0.798$ \\
    \textsc{direct}        & Yes & $0.95$ & $0.48$ & $\mathbf{0.88}$ & $64.1$ & $0.748$ \\
    \textsc{all-in-one}    & Yes & $0.93$ & $0.58$ & $\mathbf{0.88}$ & $69.8$ & $0.782$ \\
    \textbf{\textsc{two-stage}} & Yes & $0.98$ &
      $\mathbf{0.62}$ & $\mathbf{0.88}$ & $\mathbf{70.5}$ & $\mathbf{0.810}$ \\
    \bottomrule
  \end{tabular}
\end{table}

\subsection{Ablations}

We ablate each component of the pipeline. The detector-threshold sweep, the four-pass test-time augmentation, and the forged-vs-pristine training comparison were reported inline in the DTD methodology above; the contribution of language-model region filtering is quantified in Table~\ref{tab:filter}, and the effect of decomposing the reasoning across two specialized students in Table~\ref{tab:configs}. Together these results show that each design choice---thresholding at $0.40$, four-pass TTA, mixing pristine documents into DTD training, language-model region filtering, and splitting the reasoning across two students---contributes positively to the final score.

\section{Discussion}

Taken together, our experiments suggest several conclusions.

\paragraph{Reasoning supervision and decomposition are the biggest levers.}
Across every configuration, LoRA fine-tuning on distilled chains of thought dominates zero-shot prompting: for the two-stage pipeline the final score rises from $0.721$ to $0.810$ and the LLM-judge report quality jumps from $56.8$ to $70.5$, and the unstable zero-shot \textsc{all-in-one} prompt ($0.566$, with $S_\mathrm{Loc}=0.22$ and no usable explanation) is rescued to $0.782$ once trained. Teaching the model \emph{how} to reason, in the exact target format, matters more than any prompt-wording choice. Decomposition is a second lever in the same direction: given identical distilled CoTs, splitting the work across two specialized models (\textsc{two-stage}, $0.810$) beats a single model that produces the whole trace and report (\textsc{all-in-one}, $0.782$), because each model specializes and errors do not propagate across sub-tasks. A third, easily overlooked lever is the \emph{form} in which the DTD regions are presented: as our input-format comparison shows, overloading the model with extra visual input (a second heatmap image) hurts, whereas a single image annotated with numbered red boxes---paired with the same regions and coordinates in the prompt text---lets the model localize each region directly. How the evidence is presented is as consequential as what evidence is presented.

\paragraph{Filtering needs semantics and context.}
The Qwen Filterer improves localization over raw DTD regions ($\mathrm{mIoU}\;0.6402\rightarrow0.6672$), whereas purely visual filters (ResNet/U-Net) actually degrade it ($0.6113$). Deciding whether a flagged region is genuine requires reading the surrounding text, not just inspecting local pixels---which is why the pipeline first, in Stage~1, builds a small ``knowledge base'' about the document (its type, domain, language, and layout constraints) before any region is judged. This context is decisive for semantic regions in particular: a clean word substitution or an altered value is detectable only relative to what the rest of the document claims, so such a region can be confirmed or rejected only once the model knows what the document is supposed to say.

\paragraph{Dataset-driven prompt design.}
A substantial part of our effort went into studying the kinds of tampering present in the challenge data and encoding what we learned into the prompts. Three forgery families recur: \emph{Visual Clumsy} (a visible editing artefact---blur, pixelation, a redaction block, overpainting), \emph{Logical Fraud} (a value such as a number, date, ID, or name altered so that it contradicts the document's own logic, even though the glyphs render cleanly), and \emph{Semantic Subtle} (a clean word substituted for another, creating a contextual contradiction with no visual trace). We also catalogued DTD's systematic false positives---dense mathematical formulas, blank margins, and logos, stamps and seals---and turned them into explicit drop rules, so the Filterer does not blindly trust every detector region. Finally, because purely-semantic forgeries are invisible to DTD, we gave the Semantic Detective an explicit checklist to scan for: numerical inconsistencies (totals that do not match their parts, contradictory dates), impossible values (a future birth date, a wrong-length ID), internal contradictions (a name spelled two ways), semantic oxymora (``for-profit charity''), domain-impossible values (a malformed IBAN for the stated country), and unexpected mixed scripts. This taxonomy, distilled from manual inspection of the data, is what the two models' prompts and chains of thought are built around.

\section{Conclusion}

We presented Team MSU's third-place solution to the ACM MM 2026 GenText-Forensics challenge: a decomposed chain-of-thought pipeline that pairs an external document tampering detector (DTD) with two LoRA-adapted Qwen3-VL-32B models---a Filterer that validates and types detector regions, and a Semantic Detective that merges and re-grounds them, finds purely semantic anomalies, and writes the final report---both supervised by chains of thought distilled from a privileged Qwen3-VL-235B teacher. Delegating localization to a dedicated detector, grounding every finding on DTD and OCR coordinates, and splitting the reasoning across two specialized models each contributed to the result. Code and prompts are released for reproduction. During the competition our team appeared on the public leaderboard under the alias \texttt{great\_britain}.

\begin{acks}
The research was carried out using the MSU-270 supercomputer of Lomonosov Moscow State University.
We thank the GenText-Forensics challenge organizers for hosting the task and providing the RealText-V2 benchmark.
\end{acks}

\section*{Ethics and Privacy Statement}
This work develops methods for detecting and explaining forgeries in document
images, a defensive capability intended to protect administrative, financial
and medical document pipelines against manipulation. The same techniques that
locate and describe tampering could in principle inform an adversary about
which artifacts a detector is sensitive to; we mitigate this by reporting
methodology rather than attack recipes, and by using only the challenge-provided
RealText-V2 data under its research licence. 



\begin{thebibliography}{9}

\bibitem{qu2023doctamper}
Chenfan Qu, Chongyu Liu, Yuliang Liu, Xinhong Chen, Dezhi Peng, Fengjun Guo,
  and Lianwen Jin. 2023.
\newblock Towards Robust Tampered Text Detection in Document Image: New Dataset
  and New Solution.
\newblock In \emph{Proceedings of the IEEE/CVF Conference on Computer Vision
  and Pattern Recognition (CVPR)}. 5937--5946.

\bibitem{forensichub2025}
Bo Du, Xuekang Zhu, Xiaochen Ma, Chenfan Qu, Kaiwen Feng, Zhe Yang, Chi-Man
  Pun, Jian Liu, and Jizhe Zhou. 2025.
\newblock ForensicHub: A Unified Benchmark and Codebase for All-Domain Fake
  Image Detection and Localization.
\newblock \emph{arXiv preprint arXiv:2505.11003} (2025).

\bibitem{forgerygpt2024}
Fanrui Zhang, Jiawei Li, Jiaying Zhu, Esther Sun, Dong Li, Qiang Zhang, and Zheng-Jun
  Zha. 2024.
\newblock ForgeryGPT: Multimodal Large Language Model for Explainable Image
  Forgery Detection and Localization.
\newblock \emph{arXiv preprint arXiv:2410.10238} (2024).

\bibitem{logiclens2025}
Fanwei Zeng, Changtao Miao, Jing Huang, Zhiya Tan, Shutao Gong, Xiaoming Yu,
  Yang Wang, Huazhe Tan, Weibin Yao, and Jianshu Li. 2025.
\newblock LogicLens: Visual-Logical Co-Reasoning for Text-Centric Forgery Analysis.
\newblock \emph{arXiv preprint arXiv:2512.21482} (2025).

\bibitem{qwen3vl2025}
Qwen Team. 2025.
\newblock Qwen3-VL Technical Report.
\newblock \emph{arXiv preprint arXiv:2511.21631} (2025).

\bibitem{kwon2023vllm}
Woosuk Kwon, Zhuohan Li, Siyuan Zhuang, Ying Sheng, Lianmin Zheng,
  Cody Hao Yu, Joseph E. Gonzalez, Hao Zhang, and Ion Stoica. 2023.
\newblock Efficient Memory Management for Large Language Model Serving with
  PagedAttention.
\newblock \emph{arXiv preprint arXiv:2309.06180} (2023).

\bibitem{paddleocr}
PaddleOCR Team. 2025.
\newblock Introduction to PP-OCRv5.
\newblock \emph{PaddleOCR Documentation} (2025).

\bibitem{hu2022lora}
Edward J. Hu, Yelong Shen, Phillip Wallis, Zeyuan Allen-Zhu, Yuanzhi Li,
  Shean Wang, Lu Wang, and Weizhu Chen. 2022.
\newblock LoRA: Low-Rank Adaptation of Large Language Models.
\newblock In \emph{International Conference on Learning Representations (ICLR)}.

\bibitem{dettmers2023qlora}
Tim Dettmers, Artidoro Pagnoni, Ari Holtzman, and Luke Zettlemoyer. 2023.
\newblock QLoRA: Efficient Finetuning of Quantized LLMs.
\newblock In \emph{Advances in Neural Information Processing Systems
  (NeurIPS)}.

\bibitem{zhang2020bertscore}
Tianyi Zhang, Varsha Kishore, Felix Wu, Kilian Q. Weinberger, and Yoav Artzi.
  2020.
\newblock BERTScore: Evaluating Text Generation with BERT.
\newblock In \emph{International Conference on Learning Representations (ICLR)}.

\end{thebibliography}
\end{document}